\documentclass[sigconf, nonacm]{acmart}
\AtBeginDocument{%
  }

\copyrightyear{2026}
\acmYear{2026}
\setcopyright{none}
\setcctype{by}
\acmConference[RecSys '26]{20th ACM Conference on Recommender Systems}{September 27-October 02, 2026}{Minneapolis, MN, USA}
\acmBooktitle{20th ACM Conference on Recommender Systems (RecSys '26), September 27-October 02, 2026, Minneapolis, MN, USA}

\usepackage{hyperref}
\usepackage{url}
\usepackage{graphicx}
\usepackage{caption}
\usepackage{subcaption}
\usepackage{bm}
\usepackage{tabularx}
\usepackage{multirow}
\usepackage{makecell}
\usepackage{xr}
\usepackage{placeins}
\usepackage{enumitem}
\usepackage{array}
\usepackage{float}
\usepackage{dblfloatfix}
\usepackage{makecell}
\usepackage{amsmath}
\usepackage{algorithm}
\usepackage{algpseudocode}
\usepackage{tikz}
\usepackage{graphicx}
\usepackage{booktabs}

\newcommand{\commentout}[1]{%
}

\fontsize{7.8}{8}\selectfont

\begin{document}

\title{Medical Causal Hypothesis Verification \\with Large Language Models}

\author{Safiyyah Ahmed\textsuperscript{\textasteriskcentered}}
\affiliation{%
  \institution{University of Illinois Chicago}
  \city{Chicago}
  \state{IL}
  \country{USA}
 }
\email{sahme213@uic.edu}

\author{Abrar Ansari\textsuperscript{\textasteriskcentered}}
\affiliation{%
  \institution{University of Illinois Chicago}
  \city{Chicago}
  \state{IL}
  \country{USA}
 }
\email{aansa30@uic.edu}

\author{Md Aminul Islam}
\affiliation{%
  \institution{University of Illinois Chicago}
  \city{Chicago}
  \state{IL}
  \country{USA}
 }
\email{mislam34@uic.edu}

\author{Elena Zheleva}
\affiliation{%
  \institution{University of Illinois Chicago}
  \city{Chicago}
  \state{IL}
  \country{USA}
 }
\email{ezheleva@uic.edu}


\pagestyle{fancy}
\fancyhf{}
\fancyhead[LE]{\scriptsize CONSEQUENCES Workshop @ RecSys ’26, October 02, 2026, Minneapolis, MN, USA}
\fancyhead[RE]{\scriptsize Safiyyah Ahmed, Abrar Ansari, Md Aminul Islam \& Elena Zheleva}
\fancyhead[LO]{\scriptsize Medical Causal Hypothesis Verification\\ with Large Language Models}
\fancyhead[RO]{\scriptsize CONSEQUENCES Workshop @ RecSys ’26, October 02, 2026, Minneapolis, MN, USA}
\fancyfoot[C]{\scriptsize \thepage}
\renewcommand{\headrulewidth}{0pt}

\begin{abstract}
  The growing use of large language models (LLMs) for search and information retrieval underscores the need to evaluate their reliability in high-stakes domains such as healthcare. Although LLMs can effectively answer questions about diseases, symptoms, and treatments, their ability to accurately assess causal relationships and ground their conclusions in verified scientific evidence remains unclear. Here, we present a preliminary, small-scale study that investigates the accuracy of LLMs in evaluating causal medical claims and supporting them with peer-reviewed research. We propose an evaluation framework for causal hypothesis verification that can be used to systematically track the performance of existing and future LLMs. We assess the performance of eight LLMs on 17 medical causal hypotheses to evaluate whether they can reliably  verify these hypotheses using scientific evidence from the literature. We systematically annotate the scientific evidence they provide according to six criteria (a total of 1,067 annotation points) and assess them with nine evaluation metrics. Our analysis shows that while LLMs exhibit strong recall, they often perform poorly at providing valid scientific articles and evidence for support and at rejecting unsupported hypotheses. These findings highlight a critical limitation of current LLMs, as they cannot yet be trusted fully to verify causal relationships from the biomedical literature. This work underscores the need for rigorous evaluation before using LLMs for search and retrieval in healthcare settings.

\end{abstract}
\thanks{\textit{CONSEQUENCES Workshop @ RecSys ’26, October 02, 2026, Minneapolis, MN, USA}}


\maketitle

\section{Introduction}

Causal reasoning plays a central role in decision making and predicting the effects of interventions, especially in fields such as medicine, science, and public policy. 
In many real-world settings, causal claims are expressed in unstructured text, such as scientific papers, clinical notes, policy reports, news, and social media discussions. Large language models (LLMs) are particularly good at processing and organizing such unstructured text at scale and have the potential to act as assistants in finding evidence for causal hypotheses and mechanisms. However, in high-stakes environments, such as in medical and legal settings, it is important to exercise caution and to carefully assess the limitations and potential pitfalls of relying on these models.

There is a growing body of literature evaluating the capabilities of LLMs for causal inference and reasoning, broadly categorized under model reasoning, commonsense reasoning, counterfactual reasoning, and fairness and debiasing~\citep{liu-etal-2025-large-language}. Early studies have demonstrated the emerging ability of LLMs to elicit causal knowledge from text (e.g.,~\cite{kiciman2023causal}), while others have raised concerns about their parrot-like behaviors~\citep{zevcevic2023causal} and their limitations in reasoning over causal graphs (e.g.,~\cite{jin2023cladder}). However, these studies have not considered the retrieval capabilities of LLMs in providing scientific evidence to substantiate causal claims. Recent work on retrieval-augmented generation has explored the use of LLMs for verifying scientific claims more generally~\citep{wang2025llm_evidence_retrieval,mohole2025verirag}, but it has not specifically evaluated their capabilities for verifying causal claims.



In this work, we conduct a small-scale study on causal hypothesis verification with the goal of evaluating whether LLMs can be used as assistants in answering causal queries and finding evidence from the medical literature to support their answers. We design a prompt that asks an LLM to verify a specific medical causal hypothesis of the form "diabetes causes renal disease" and to provide up to three scientific articles if the hypothesis is supported. While the number of hypothesis we evaluate is fairly small (17), it is important to note that 1) when this study was conducted, these hypotheses were not available on the web and therefore could not have been used by the LLMs in their training, 2) the hypotheses have been validated by clinicians, and 3) we annotate all the LLM answers across multiple dimensions for a total of $1,067$ annotation points which constitutes a significant annotation effort. We propose an evaluation framework for causal hypothesis verification that can be used to systematically track the performance of existing
and future LLMs. Our results highlight some of the limitations with using LLMs as assistants in health-related decision making scenarios. 

\begingroup
\renewcommand{\thefootnote}{}
\footnotetext{\textsuperscript{\textasteriskcentered}Both authors contributed equally to this research.}
\endgroup

\section{Evaluation Setup for Causal Hypothesis Verification }



\begin{figure*} [b]
\vspace{-1em}
\centering
\captionsetup{justification=raggedright, margin=0cm}
\setlength{\fboxsep}{6pt}
\setlength{\fboxrule}{0.8pt}
\fbox{%
\begin{minipage}{0.96\textwidth}
\footnotesize\ttfamily
Imagine you are a medical researcher and you are trying to provide evidence for the hypothesis that [hypothesis]. If such evidence exists from the peer-reviewed scientific literature, specify the three scientific articles that provide the strongest evidence for this hypothesis in a specific format I will provide. For each scientific article you find, provide the exact URL to the DOI record, the exact paper title, the exact paper abstract as written by the authors, and a quote from the article that specifically points to the support for the hypothesis. Format your response like so:

"Yes" or "No" (answer one) only once at the very top of your response whether such evidence exists

If the evidence exists, then for each scientific article include:

"URL: "

"Paper title: "

"Paper Abstract: "

"Quote from article supporting hypothesis: "

Emphasis on EXACT. I don’t want summaries of abstracts, summaries of quotes or summaries of paper titles.

Don't include those terms until the final response. (Don't include the words "URL", "title", "Abstract", "Quote" in the reasoning itself, only in the final response)

In front of each URL, type URL: and include the actual link not a hyperlink. In front of each paper title write "Paper Title: " and in front of each abstract write "Abstract: ". In front of each quote write "Quote: " Include nothing else in your response.

Prioritize high-impact human studies, emphasize both mechanistic and clinical outcome data, and include both open-access and paywalled articles when necessary.

If no evidence exists, please explain the process by which you have completed the search to conclude that no such articles exist.
\end{minipage}%
}
\caption{Standardized prompt template used for all LLMs and all hypotheses.}
\label{fig:prompt}
\end{figure*}

To evaluate whether LLMs can reliably provide evidence for causal relationships, we design an automated pipeline that uses a structured prompt to collect and parse responses from LLMs. The responses are annotated by humans for accuracy and then thoroughly evaluated. We design the prompt and evaluation metrics guided by the following two research questions:
\begin{itemize}[leftmargin=10pt, nosep]
    \item \textbf{RQ1:} Can LLMs accurately determine the validity of medical causal hypotheses?
    \item \textbf{RQ2:} Do LLMs produce accurate and relevant scientific evidence to support their causal claims?
\end{itemize}

We use a set of 17 causal hypotheses discovered in prior work \citep{adhikari-arXiv25}, which analyzes ICD-10 diagnosis codes associated with repeat emergency room (ER) visits among diabetic patients. The hypotheses have been validated by four expert clinicians, therefore we have ground truth for their validity. The hypotheses include relationships between comorbid conditions, such as diabetes, hypertension, renal disease, obesity, and edema, and their potential causal pathways influencing repeat ER utilization. 
Considering a mixture of  supported and unsupported claims (11 of the 17 hypotheses are supported) ensures that the evaluation remains meaningful, while also enabling an effective assessment of an LLM’s ability to evaluate causal claims. Hypotheses are in the form "chronic peripheral venous insufficiency causes chronic non-pressure foot ulcers."


\textbf{Prompt construction}. We develop a standardized prompt template (Figure~\ref{fig:prompt}) that is applied consistently across all LLMs and hypotheses. The prompt explicitly asks whether the evidence supports the hypothesis. If support is indicated, the model is instructed to provide the three strongest pieces of evidence from peer-reviewed scientific papers. For each, it must include the DOI link, the exact article title, the full abstract of the paper, and any direct quotes from the article supporting the hypothesis. This design ensures highly structured outputs, facilitating automated parsing and systematic evaluation. 
If the scientific literature does not support the causal hypothesis, the model is instructed to respond with “No” and to explain how it reached this conclusion. These specific response components are selected because they can provide the essential information needed to verify factual accuracy within the scientific literature. The ability to return real articles, valid DOI links, accurate abstracts, and relevant quotes is important for assessing whether the LLM is grounding its reasoning in evidence-based sources rather than generating  hallucinated claims. 

\textbf{Evaluated LLMs and implementation details.} We use eight LLMs for causal hypothesis verification, including \textbf{Mistral}, \textbf{Deepseek-R1}, \textbf{Deepseek-V3}, \textbf{Gemini Flash}, \textbf{Gemini Pro}, \textbf{Qwen}, \textbf{GPT-4o}, and \textbf{Llama-4-Maverick}. We select these LLMs based on the availability of API access and free-tier usage limits, as our study requires programmatic execution of all prompts at scale and automatic storage of results for downstream evaluation. The only exception is GPT-4o, which does not offer a free tier, and from which we collected responses manually. We use these models based on the most recent versions available at the time of data collection. We provide the constructed prompt to each LLM for each causal hypothesis and collect its response. Then we automatically parse it according to the prompt instructions. 
As a result, 
we obtain responses from all eight LLMs across $17$ hypotheses, for a total of $136$ responses with $3$ articles for each response that supports the hypothesis. 
In rare cases where automatic parsing fails, we manually extract the components from the response. 

\textbf{Hypothesis and supporting evidence annotation}. Once we obtain responses from the LLMs, we manually verify the validity of their outputs at both the hypothesis and evidence levels. First, we assess whether the model’s decision regarding the hypothesis is correct, independent of any evidence it provides, by determining whether the response matches the ground truth obtained from clinicians \textit{(Correct Conclusion)}. If the LLM claims that the hypothesis is supported, we then annotate the validity of its supporting evidence. Specifically, each provided article is annotated according to whether an article with the stated title exists \textit{(Real Paper)} and whether the DOI link correctly corresponds to the provided article title \textit{(Correct URL)}. If an article is a \textit{Real Paper}, then we also annotate whether the abstract is provided accurately from the original paper \textit{(Real Abstract)}, whether the quoted text appears exactly in the paper and supports the model’s stated causal claim \textit{(Real Quote)}, whether the article itself supports the hypothesis \textit{(Paper Supporting Hypothesis)}. For the papers that do not support the hypothesis, we annotate whether the article reports only an association rather than a direct causal relationship \textit{(Association)}. For the real papers, if the DOI URL is incorrect, we also find the correct URL. 
This multi-stage evaluation is essential because an LLM may correctly identify a hypothesis as true while simultaneously hallucinating or misrepresenting article metadata, references, or textual evidence. To facilitate annotation, we use Label Studio \citep{LabelStudio}, which enables structured and efficient annotation of each component. 


To ensure reliability, two annotators independently label the entire set of LLM responses. Then they compare their annotations and discuss any disagreements. In cases of disagreement, the responses are re-examined to verify their validity and reach a final consensus. We use the final agreed-upon annotated dataset for all our evaluations. The final set contains $1,067$ annotation points covering whether each provided paper is real, and if so, answering the five criteria for each real paper.

\textbf{Evaluation metrics.} We evaluate LLM performance using precision, recall, F1 score, specificity, and accuracy for accurate hypothesis classification. Precision measures the proportion of an LLM’s positive classifications that correspond to actual causal relationships, while recall measures the proportion of all true causal relationships correctly identified by the LLM. Specificity measures the proportion of false hypotheses correctly identified, and the F1 score provides the harmonic mean of precision and recall. We also measure the accuracy of the evidence provided by LLMs when classifying hypotheses as true, considering the proportion of real or valid papers out of the total number of papers. Similarly, we evaluate the accuracy of URLs or DOI links, as well as the correctness of abstracts and direct quotes provided by the LLMs. 

\section{Results}
\begin{figure} [b]
\vspace{-1.0em}
    \centering
    \captionsetup{justification=raggedright, margin=0cm}
    \includegraphics[width=\columnwidth]{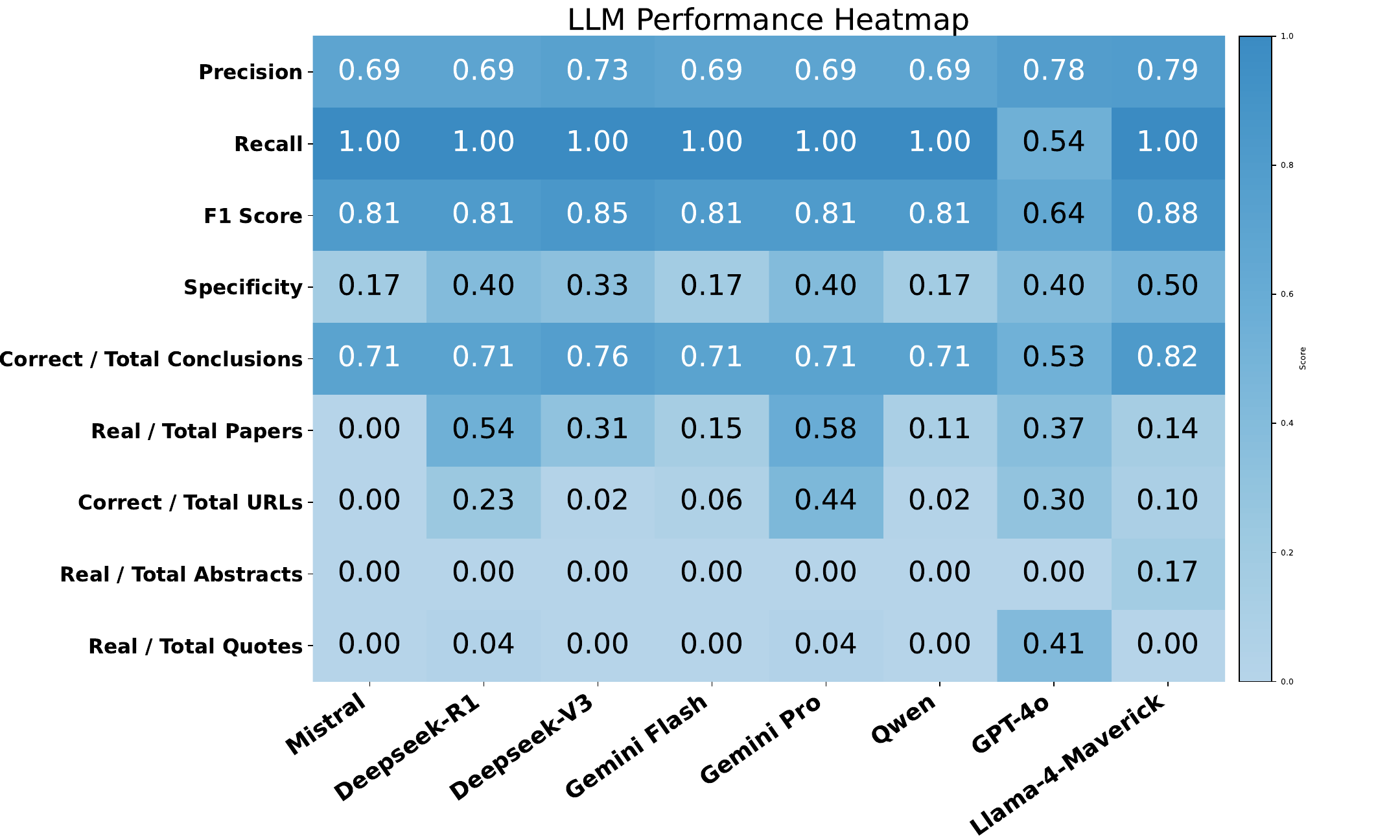}
    \captionsetup{width=\columnwidth}
    \caption{LLM performance across accuracy and evidence-grounding metrics.}
    \label{fig:llm-heatmap}
    \vspace{-1em}
\end{figure}

In this section, we evaluate the performance of LLMs with respect to our research questions. The main results are reported in Figure~\ref{fig:llm-heatmap}.

\textbf{Causal hypothesis classification performance (RQ1).}
The LLMs achieve accuracy (correct/total conclusions) between $0.53$ and $0.82$ and F1 scores ranging from $0.64$ to $0.88$, reflecting moderate overall performance in classifying causal hypotheses. Precision values in the range of 0.69-0.79 indicate a tendency to over-predict true causal hypotheses, resulting in a higher number of false positives. 
A recall of $1.0$ for most models indicates that these models correctly identified all true causal hypotheses, whereas GPT-4o (recall = $0.54$) failed to detect a substantial portion of true causal relationships. 
The low specificity observed for all models ($0.17$–$0.50$) indicates a limited capacity to correctly identify false causal hypotheses, resulting in a high false positive rate. 
Overall, the results suggest that LLMs are generally more effective at identifying true causal hypotheses than rejecting false ones, achieving high recall but relatively low specificity, which leads to a tendency toward false positive predictions.

\begin{table*}
\setlength{\tabcolsep}{3.0pt} 
\fontsize{7.8}{9}\selectfont
\captionsetup{justification=raggedright, width=1.0\textwidth}
\caption{Number of real or valid evidences and total evidences provided by LLMs.}
\vspace{-0.1cm}
\label{tab:results}
\centering
\begin{tabular}{ccccccccc}
\toprule
\textbf{Metric} & \textbf{Mistral} & \textbf{Deepseek-R1} & \textbf{Deepseek-V3} & \textbf{Gemini Flash} & \textbf{Gemini Pro} & \textbf{Qwen} & \textbf{GPT-4o} & \textbf{Llama-4-Maverick} \\
\midrule
Correct conclusion of hypothesis & 12 & 12 & 13 & 12 & 12 & 12 & 9 & 14 \\
Real papers & 0 & 26 & 14 & 7 & 28 & 5 & 17 & 6 \\
Number of papers supporting hypothesis & 0 & 18 & 11 & 4 & 21 & 3 & 9 & 1 \\
Total papers & 46 & 48 & 45 & 48 & 48 & 46 & 46 & 42 \\
Correct URLs & 0 & 11 & 1 & 3 & 21 & 1 & 14 & 4 \\
Total URLs & 46 & 48 & 45 & 48 & 48 & 46 & 46 & 42 \\
Real quotes & 0 & 1 & 0 & 0 & 1 & 0 & 7 & 0 \\
Total quotes & 0 & 26 & 14 & 7 & 28 & 5 & 17 & 6 \\
Real abstract & 0 & 0 & 0 & 0 & 0 & 0 & 0 & 1 \\
Association count & 0 & 1 & 1 & 0 & 2 & 0 & 3 & 0 \\
\bottomrule
\end{tabular}
\end{table*}

\textbf{Evidence accuracy and scientific grounding (RQ2).}
In RQ2, we evaluate whether hypotheses classified as true by LLMs are supported by scientific evidence. This evaluation measures the reliability of LLM decisions with respect to scientific sources, since a model may correctly label a hypothesis as true while hallucinating or misrepresenting article metadata. 
LLMs often fail to provide verifiable scientific papers to support their claims with percent valid papers ranging from $11\%$ (Qwen) to $58\%$ (Gemini Pro). The percent correct URLs provided by the LLMs is also low, ranging from $0\%$ (Mistral) to $44\%$ (Gemini Pro). This indicates that, even when papers are provided, LLMs often fail to supply correct sources, either due to hallucinated citations or an inability to accurately extract source information for real papers. 
All LLMs fail to accurately extract exact abstracts from the cited papers, except for Llama-4-Maverick, which achieves an accuracy of 17\%. Similarly, all LLMs rarely provide correct exact quotes from papers supporting the hypotheses ($0$--$4\%$), except GPT-4o, which achieves 41\% accuracy.  
It implies that all LLMs struggle to reliably ground their causal claims in verifiable evidence from scientific papers, whereas GPT-4o demonstrates a stronger ability to extract such evidence from source documents. 

Table~\ref{tab:results} shows the raw number of real or valid pieces of evidence provided by each LLM, along with the total number of evidence pieces. Across models, LLMs produced a similar number of correct conclusions for hypotheses classification, ranging from 9 to 14 out of 17. However, there is substantial variation in the verifiability of supporting evidence provided by LLMs. DeepSeek-R1 and Gemini Pro yield a comparatively high number of real papers (26 and 28, respectively) as well as correct URLs (11 and 21), whereas models such as Qwen and Mistral produce few or no real papers or valid URLs. Although most models generate a comparable number of total papers and URLs (roughly 42--48), the proportion of valid evidence differs substantially across models, indicating that generating many references does not necessarily correspond to providing reliable supporting evidence. The number of papers that actually support the hypothesis also varies considerably, with Gemini Pro and DeepSeek-R1 providing the highest counts (21 and 18, respectively), while several models produce very few supporting papers. Real quotes supporting the hypotheses are generally rare (ranging from 0 to 7), with GPT-4o being a slight exception, producing seven instances. In contrast, real abstracts generated by the LLMs are extremely limited, appearing only once for Llama-4-Maverick. The association count, which indicates whether an article reports only an association rather than a direct causal relationship, is low across all models (0--3), showing that only a small portion of the evidence provided by LLMs reports associations instead of potential causal relations. Overall, all of these results suggest that although LLMs often reach correct conclusions, the supporting evidence they provide such as papers, URLs, quotes, and abstracts is highly variable. In many cases, this evidence is not reliable, highlighting the need for careful verification when using LLM-generated references for causal hypothesis verification.

\begin{table*}
\setlength{\tabcolsep}{5.0pt} 
\fontsize{7.8}{9}\selectfont
\captionsetup{justification=raggedright, width=1.0\textwidth}
\caption{RAG architecture usage and model size of evaluated LLMs.}
\vspace{-0.1cm}
\label{tab:model_details}
\centering
\begin{tabular}{ccccccccc}
\toprule
\textbf{Metric} & \textbf{Mistral} & \textbf{Deepseek-R1} & \textbf{Deepseek-V3} & \textbf{Gemini Flash} & \textbf{Gemini Pro} & \textbf{Qwen} & \textbf{GPT-4o} & \textbf{Llama-4-Maverick} \\
\midrule
RAG architecture usage
& No & No & No & No & Yes & No & Yes & Yes \\

Model size 
& 12B & 671B / 37B active & 671B / 37B active & 60--300B 
& 128--140B & 30B & 1.7T & 400B / 17B active \\
\bottomrule
\end{tabular}
\vspace{-1em}
\end{table*}

\textbf{LLM search process.} Our prompt asks each LLM to explain how it decided when there is no support for a hypothesis. Deepseek R1, ChatGPT, and Gemini Pro demonstrate more sophisticated reasoning, while other models, including Mistral, Qwen, and Deepseek V3 rely more on surface level keyword searches for papers directly mentioning the hypothesis. Gemini Flash and Llama-4-Maverick have a mix of the two approaches. Their performance reflects the level of reasoning. 
Deepseek R1 uses search using MeSH terms and keywords and then checking whether any of the articles establish causal relationships between the specified diseases. It rejects as evidence any articles that show mere associations. 
ChatGPT's five-step approach: defining the hypothesis, searching databases, applying selection criteria, reviewing high-impact studies, and distinguishing between association and causation, reflects a structured methodology that is anchored in real, identifiable sources. 
Gemini-pro's process involves combining hypothesis keywords like "hyperlipidemia" and "vaccination" with terms like "predicts" and "determinants," then taking the extra step of classifying what it finds into three distinct thematic buckets, and finally explaining why even the most relevant bucket, the association between hyperlipidemia and vaccination rates doesn't support causation due to confounding variables like age, comorbidities, and healthcare utilization. 
Gemini Flash improves on basic keyword search by expanding to related terms like "patient motivation for general examination due to hyperlipidemia" and "reasons for adult physical exam hyperlipidemia" and incorporating some causal reasoning, producing balanced but slightly less nuanced outputs. 
Llama’s reasoning follows a clean two-phase structure: an initial broad search, followed by a refined second search with tighter filters, paying attention to the biology underlying the hypothesis. 
Deepseek V3 primarily relies on searching through databases like PubMed/MEDLINE and Google Scholar and using keywords such as "hyperlipidemia", "etiology", "risk factor", "general medical examination", "health check-up", "periodic health examination", and filtering out papers that do not explicitly prove causation. 
Mistral also does a keyword search through databases like PubMed, Web of Science, and Google Scholar, but mainly provides illustrative examples of papers without strong causal analysis. 
Qwen names databases, lists explicit keyword strings, and jumps straight to a conclusion without unpacking its reasoning. 


\textbf{Model size effect on responses. } As shown in Table~\ref{tab:model_details}, Mistral and Qwen are the smallest models with around 12 billion and 30 billion parameters respectively which likely explains their low specificity ($0.17$) and low percentage of real retrieved papers. 
In contrast, larger models generally performed better. Gemini 2.5 Pro, with an estimated $128-140$ billion parameters, achieves the highest values for real papers ($0.58$) and correct URLs ($0.44$), while DeepSeek-R1, despite using only $37$ billion active parameters per token, also performs well with $0.54$ real papers. GPT-4o has the highest number of real quotations ($0.41$), and Llama-4-Maverick achieved the highest F1 score ($0.88$), specificity ($0.50$), and correct conclusions ($0.82$). However, model size alone does not fully explain performance. Although the larger models generally outperform the smaller ones, they excel in different evaluation metrics, suggesting that model architecture, training data, and retrieval capabilities contributed as much to performance as parameter count. 

\textbf{RAG architecture effect on responses.} We ask each LLM if they retrieve any external documents to answer the prompt. As shown in Table~\ref{tab:model_details}, Mistral, DeepSeek-R1, DeepSeek-V3, and Qwen stated that they do not retrieve external documents or search the web and their responses are generated from their internal knowledge. In contrast, GPT-4o states that it retrieves and analyzes external scientific documents and bibliographic records, while Llama-4-Maverick reports retrieving several external documents using various databases. Gemini Flash and Gemini Pro state that they could retrieve information from the web, but Gemini Flash state that it does not do so for the prompt, whereas Gemini Pro stated that it could provide titles, DOI links, and summaries of scientific literature. Models that are retrieving external information generally produced better results, including higher numbers of real papers, valid URLs, and accurate quotations, versus models that relied exclusively on internal knowledge. However, document retrieval does not guarantee the best overall performance, as Llama-4 Maverick achieves the highest F1 score (0.88) and conclusion accuracy (0.82) despite emphasizing internal reasoning rather than active document retrieval. This suggests that both external document retrieval and the model's internal reasoning abilities contributed to the quality and verifiability of scientific evidence generation.

\section{Conclusion}
This work provides a preliminary study that evaluates the capacity of different LLMs to do causal hypothesis verification in the biomedical settings, assessing their ability to judge causal validity and to ground conclusions in verifiable scientific 
evidence. LLMs sometimes perform well at identifying true causal relationships, but they consistently struggle to reject unsupported claims and to reliably reference real, traceable evidence from the scientific literature, often producing hallucinated citations. These findings highlight a gap between causal reasoning and trustworthy scientific evidence grounding and should be further investigated on a larger scale. These limitations indicate that, in their current form, LLMs should be used with caution for causal verification tasks and should not be relied upon as autonomous tools for evidence-based biomedical reasoning.


\bibliographystyle{ACM-Reference-Format}
\bibliography{main}


\end{document}